%% file: main.tex
\documentclass{article}

\usepackage{arydshln}
\usepackage{microtype}
\usepackage{graphicx}
\usepackage{booktabs} %
\usepackage{makecell}
\usepackage{soul}
\usepackage{float}
\usepackage{adjustbox}
\usepackage[math]{cellspace}
\usepackage{multirow}
\usepackage{pifont}
\usepackage{caption}
\usepackage{subcaption}
\usepackage[dvipsnames, svgnames, x11names]{xcolor}

\usepackage{lipsum}

\usepackage{hyperref}

\newcommand{\xmark}{{\color{red}\ding{55}}}%

 \usepackage[preprint]{neurips_2025}

\usepackage{amsmath}
\usepackage{amssymb}
\usepackage{mathtools}
\usepackage{amsthm}

\DeclareMathOperator*{\argmin}{arg\,min}

\usepackage[capitalize,noabbrev]{cleveref}

\theoremstyle{plain}

\theoremstyle{definition}

\theoremstyle{remark}

\usepackage[
  colorinlistoftodos,    %
  textsize=tiny,         %
  disable,               %
]{todonotes}
\setuptodonotes{color=blue!30}

\title{Semi-Explicit Neural DAEs: Learning Long-Horizon Dynamical Systems with Algebraic Constraints}

\author{%
  Avik Pal\\
  MIT CSAIL\\
  \texttt{avikpal@mit.edu}
  \And
  Alan Edelman\\
  MIT CSAIL\\
  \texttt{edelman@mit.edu}
  \And
  Chris Rackauckas\\
  MIT CSAIL\\
  \texttt{crackauc@mit.edu}
}

\begin{document}

\maketitle

\begin{abstract}

Despite the promise of scientific machine learning (SciML) in combining data-driven techniques with mechanistic modeling, existing approaches for incorporating hard constraints in neural differential equations (NDEs) face significant limitations. Scalability issues and poor numerical properties prevent these neural models from being used for modeling physical systems with complicated conservation laws. We propose Manifold-Projected Neural ODEs (PNODEs), a method that explicitly enforces algebraic constraints by projecting each ODE step onto the constraint manifold. This framework arises naturally from semi-explicit differential-algebraic equations (DAEs), and includes both a robust iterative variant and a fast approximation requiring a single Jacobian factorization. We further demonstrate that prior works on relaxation methods are special cases of our approach. PNODEs consistently outperform baselines across six benchmark problems achieving a mean constraint violation error below $10^{-10}$. Additionally, PNODEs consistently achieve lower runtime compared to other methods for a given level of error tolerance. These results show that constraint projection offers a simple strategy for learning physically consistent long-horizon dynamics.

\end{abstract}

\begin{figure}[t]
    \centering
    \includegraphics[width=0.94\textwidth]{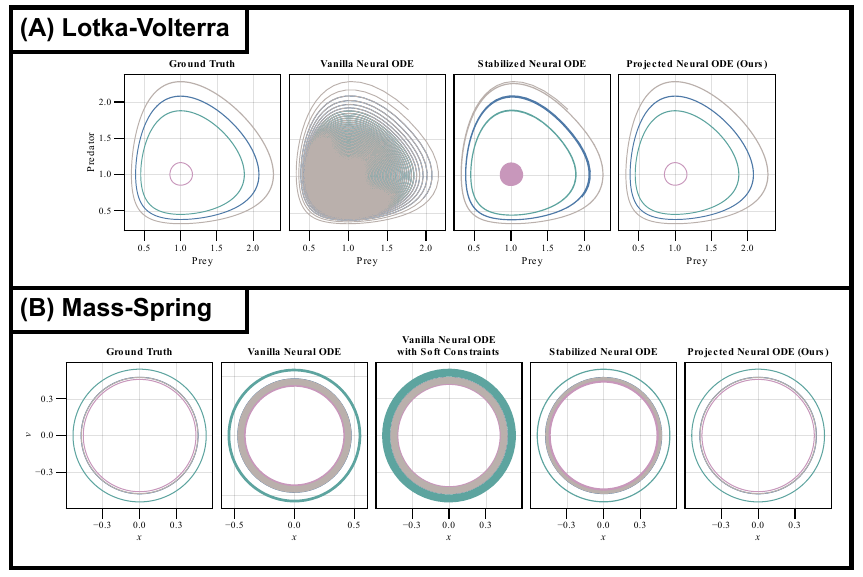}
    \caption{\textbf{Qualitative comparison of learned dynamics across neural ODE variants.}   We show predicted trajectories from various neural ODE models on two benchmark problems: (a) Lotka-Volterra predator-prey dynamics and (b) a 2D mass-spring system with conserved energy. Vanilla neural ODEs diverge over time due to lack of constraint enforcement. Adding soft penalties improves short-term behavior but fails to prevent long-term drift. Stabilized NODEs (SNODEs) reduce drift but still show degradation over extended horizons. In contrast, PNODEs stay aligned with the ground truth throughout, accurately preserving both dynamics and algebraic invariants. These results highlight our key insight: \textit{explicit projection onto the constraint manifold enables qualitatively superior and physically consistent learning.}}\label{fig:motivation-figure}
    \vspace{-1em}
\end{figure}

\section{Introduction}\label{sec:introduction}

SciML has shown promise in mixing data-driven techniques with mechanistic modeling to improve the scalability of solving essential problems such as inverse problems, uncertainty quantification, and optimal control. Integrating hard constraints in SciML models like neural differential equations is a critical challenge with far-reaching implications across various scientific and engineering domains. These constraints arise from conservation laws (e.g., energy, momentum), physical symmetries, or kinematic relationships in mechanical systems. Learning models that capture both the dynamics and these invariants is essential for generating stable, long-horizon predictions -- particularly in safety-critical applications such as robotics, fluid simulation, and multi-body dynamics.

Neural Ordinary Differential Equations (ODEs)~\citep{chen2018neural} provide a flexible, data-driven framework for modeling continuous-time dynamics. However, they assume fully unconstrained evolution and offer no built-in mechanism to enforce physical invariants. In practice, learned trajectories from vanilla neural ODEs often drift away from the true manifold over time, leading to instability and non-physical behavior. Existing approaches attempt to address this through soft constraint penalties~\citep{raissi2019physics}, structure-aware architectures such as Hamiltonian or Lagrangian Neural Networks~\citep{greydanus2019hamiltonian, cranmer2020lagrangian}, or stabilization terms~\citep{white2023stabilized} that nudge trajectories back toward the constraint surface. However, these methods either fail to guarantee constraint satisfaction, require problem-specific designs, or introduce significant inference-time overhead. As a result, reliable modeling of constrained dynamics remains an open challenge in scientific machine learning.

In this work, we propose Projected Neural ODEs (PNODEs), a principled approach to learning constrained dynamics by explicitly enforcing algebraic constraints at every integration step. Our method draws inspiration from classical numerical solvers for semi-explicit differential-algebraic equations (DAEs), which maintain constraint satisfaction by projecting each step onto the valid manifold. PNODEs incorporate this projection directly into the ODE solver, ensuring that all model outputs satisfy the constraint equations exactly. We introduce two variants of PNODEs: a robust projection method that solves a nonlinear system at each step, and a fast approximation that performs a single Jacobian factorization. Both variants are compatible with gradient-based learning via implicit differentiation, and unify prior relaxation-based methods as special cases.

To evaluate our method, we benchmark PNODEs on six dynamical systems with known constraints. Across all systems, PNODEs significantly outperform baselines in both constraint violation and state prediction. For instance, our models achieve constraint errors below $10^{-10}$ on average, often improving over Stabilized Neural ODEs by several orders of magnitude, while also reducing inference time at equivalent accuracy. Our results demonstrate that explicit projection is a simple and general tool for enforcing structure in learned dynamical systems. By bridging ideas from classical numerical analysis and modern neural modeling, PNODEs offer a simple and elegant solution for learning stable, physically consistent dynamics over long horizons.

\section{Background}\label{sec:background}

\paragraph{Differential Algebraic Equations}

DAEs describe systems where dynamics evolve subject to both differential equations and algebraic constraints~\citep{von1999multibody, griepentrog1986differential, hairer2006numerical, hairer2011solving}. These algebraic constraints are dependent variables whose derivatives don't appear in the equations. A special but important sub-category under DAEs are semi-explicit DAEs or ODEs with constraints:
\begin{equation}
    \frac{\mathrm{d}u}{\mathrm{d}t} = f(u, \theta, t) \qquad g(u, t) = 0
\end{equation}
Here, $f$ governs the dynamics and $g$ enforces constraints. DAEs naturally arise in systems with conservation laws, kinematic linkages, or implicit physical laws -- common in physics, biology, and engineering. Unlike standard ODEs, DAEs require specialized numerical solvers to ensure trajectories stay on the constraint manifold. Violating these constraints, even slightly, can destabilize the solution or lead to non-physical behavior.

\paragraph{Neural Differential Equations}

Neural ODEs~\citep{chen2018neural, rackauckas2020universal, kidger2022neural} learn continuous-time dynamics by parameterizing $f(u, \theta, t)$ with a neural network and using a numerical ODE solver for the forward simulation. Given initial conditions and observations, parameters $\theta$ are optimized to minimize a loss between the model-predicted and true trajectories. While flexible and efficient, vanilla neural ODEs assume fully unconstrained dynamics and offer no built-in mechanism to enforce algebraic conditions.

\paragraph{Enforcing Constraints in Neural Differential Equations}

Two common approaches for embedding constraints in neural differential equations~\citep{karniadakis2021physics} are: (a) incorporating constraints via architectural modifications, and (b) adding soft constraint penalties to the loss function. Lagrangian Neural Networks (LNNs)\citep{cranmer2020lagrangian} and Hamiltonian Neural Networks (HNNs)\citep{greydanus2019hamiltonian} encode physics-informed priors by preserving the underlying structure of dynamical systems. Prior work~\citep{david2023symplectic,mattheakis2022hamiltonian} has demonstrated their applicability to various physics-based learning tasks. However, \citet{gruver2022deconstructing} show that most of the performance gains in these models arise from the second-order structure of the network, rather than from enforcing Hamiltonian dynamics directly.

\begin{figure}
    \centering
    \includegraphics[width=.9\linewidth]{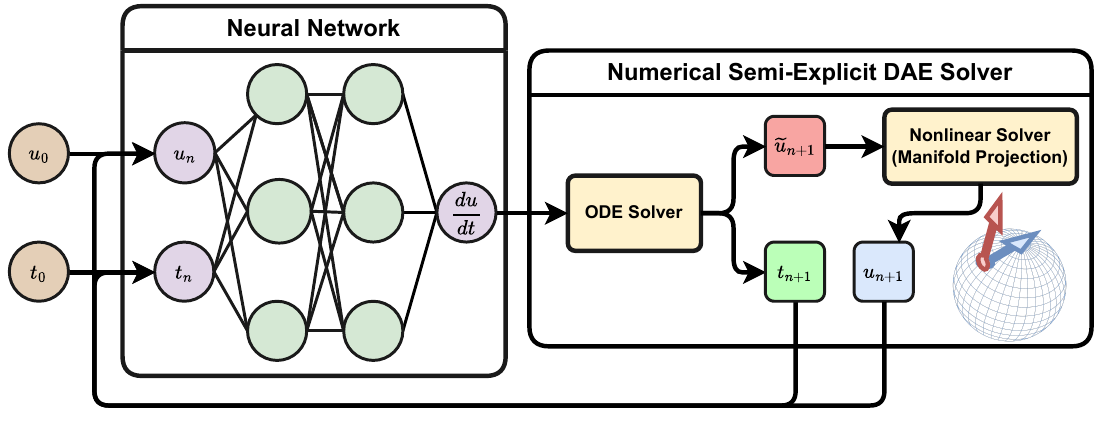}
    \caption{\textbf{Overview of the Projected Neural ODE Architecture.} Our model augments a standard Neural ODE solver by adding a constraint projection step. At each integration step, the neural network predicts an unconstrained update, which is then projected back onto the constraint manifold using a nonlinear solver. This projection enforces algebraic constraints explicitly, ensuring the system evolves on the constraint manifold. The result is a semi-explicit neural DAE framework that combines data-driven dynamics with rigorous constraint satisfaction.}\label{fig:model-overview}
\end{figure}

Another widely-used strategy is to penalize constraint violations through the loss function, as popularized by PINNs~\citep{raissi2019physics}. For instance, \citet{lim2022unifying} construct task-specific penalty terms to enforce inductive biases, but their method requires problem-specific tuning. Neural DAEs~\citep{koch2024neural} decompose the system into an ODE component and a physics-based algebraic constraint. The physics-based algebraic sub-component is trained with a PINN-inspired loss function. In our experiments, we demonstrate that using penalty loss functions doesn't lead to long-horizon stable trajectories. Additionally, as demonstrated by our experiments, certain constraints can destabilize the training with soft constraints~\citep{rathore2024challenges}.

Stabilized Neural ODEs~\citep{white2023stabilized} introduce corrective forces to keep trajectories near the constraint manifold. While broadly applicable and compatible with standard solvers, they incur significant inference-time overhead and may still allow constraint drift. Alternatively, \citet{kasim2022constants} attempt to discover and enforce latent constraints, rather than working with known constraint structures. \citet{white2024projectedneuraldifferentialequations} apply projection-based methods to enforce constraints, which closely aligns with the core idea of our work. However, their formulation is a special case of our method and is trained using continuous adjoints. This can introduce instabilities, as interpolating between two projected points does not guarantee staying on the constraint manifold unless a geodesic path is used -- leading to potential errors during backpropagation. In contrast, our framework supports both continuous and discrete adjoints and applies broadly to all projection-based updates.

\section{Solving Manifold Projected Neural ODEs as a Semi-Explicit DAE}\label{sec:method}

Many physical systems evolve under algebraic constraints that restrict the allowable states of the system to a manifold. These constraints are often known in closed form and arise from conservation laws, kinematic relations, or geometric structure. To model such systems, we build on the classical theory of DAEs, and in particular, semi-explicit DAEs that separate differential and algebraic components. This section introduces our formulation of PNODEs, which incorporate explicit constraint enforcement into neural ODE solvers by projecting each integration step onto the constraint manifold. \Cref{fig:model-overview} summarizes our training pipeline and solver architecture. At each ODE step, the model first predicts an unconstrained update using a neural network, followed by a projection onto the constraint manifold using either a robust iterative solver or a simplified method with a fixed Jacobian. During training, gradients are backpropagated through both the solver and projection using discrete adjoints and the implicit function theorem. This approach allows us to explicitly enforce constraints while maintaining end-to-end differentiability and efficiency.

\paragraph{Solving a Semi-Explicit DAE}

We are solving a system where $f(u, \theta, t)$ is parameterized by a neural network. Our manifold $\mathcal{M}$ is given by the set $\left\{u: g(u, t) = 0\right\}$. This forms the following semi-explicit DAE:
\begin{equation}
    \frac{\mathrm{d}u}{\mathrm{d}t} = f(u, \theta, t) \qquad g(u, t) = 0
\end{equation}
For $u_n \in \mathcal{M}$, the distance of $\tilde{u}_{n + 1}$ to the manifold $\mathcal{M}$ is the size of the local error $\mathcal{O}\left(h^{p + 1}\right)$. We compute the projection $u_{n + 1}$ on the manifold $\mathcal{M}$ by solving the following constrained optimization problem~\citep{hairer2006geometric}:
\begin{align}
    & u_{n + 1} \in \argmin_{z} \| z - \tilde{u}_{n + 1} \|_2\\
    \text{subject to } & ~ g(z) = 0
\end{align}
Using Lagrange multipliers, we obtain the following nonlinear equations:
\begin{equation}
    \begin{aligned}
        z &= \tilde{u}_{n + 1} + \left(\frac{\partial g(z)}{\partial u}\right)^T \lambda\label{eq:nonlinear-eq-1}\\
        g(z) &= 0
    \end{aligned}
\end{equation}
This set of equations constitutes our robust variant of Manifold-Projected Neural ODE, which potentially uses multiple Jacobian factorizations to solve the nonlinear system.

\paragraph{Simplifying Assumption Requiring Single Jacobian Factorization} %

We can make a simplifying assumption by replacing $\frac{\partial g(u)}{\partial u}$ with $\frac{\partial g(\tilde{u}_{n + 1})}{\partial u}$ in \Cref{eq:nonlinear-eq-1}~\citep{hairer2006geometric}. Then, our nonlinear equation becomes:
\begin{equation}
    \begin{aligned}
        g\left(\tilde{u}_{n + 1} + \left(\frac{\partial g(\tilde{u}_{n + 1})}{\partial u}\right)^T \lambda\right) = 0
    \end{aligned}
\end{equation}
with $\lambda_0 = 0$. Additionally, we can solve this nonlinear equation with Newton's method with a fixed Jacobian~\citep{cohen1996cvode}.
\begin{equation}
    \begin{aligned}
        \frac{\partial g(\tilde{u}_{n + 1})}{\partial u} \left(\frac{\partial g(\tilde{u}_{n + 1})}{\partial u}\right)^T \delta \lambda_k &= - g\left( \tilde{u}_{n + 1} + \left(\frac{\partial g(\tilde{u}_{n + 1})}{\partial u}\right)^T \lambda_{k} \right)\\
        \textbf{}\lambda_{k + 1} &= \lambda_k + \delta \lambda_k
    \end{aligned}
\end{equation}

We refer to this model as the Manifold-Projected Neural ODE . While our experiments suggest that this variant is a faster version, there are cases where this model doesn't show promising outcomes and hence the robust variant should be preferred in those cases.

\paragraph{Stabilized NODE as a Special Case of Manifold Projected NODE} %

Now consider a single step of this method and assume convergence to the solution in a single step:
\begin{equation}
    u_{n + 1} = \tilde{u}_{n + 1} - \mathcal{J}_{u}^T \left(\mathcal{J}_{u} \mathcal{J}_{u}^T\right)^{-1} g\left( \tilde{u}_{n + 1}\right)
\end{equation}
where $\mathcal{J}_u^T = \frac{\partial g(\tilde{u}_{n + 1}))}{\partial u}$. Consider Forward Euler's method where $\tilde{u}_{n + 1}$ is computed as:
\begin{equation}
    \begin{aligned}
        & \tilde{u}_{n + 1} = u_n + h f(u_{n}, \theta, t_n) \\
        \implies & u_{n + 1} = u_n + h f(u_{n}, \theta, t_n) - \mathcal{J}_{u}^T \left(\mathcal{J}_{u} \mathcal{J}_{u}^T\right)^{-1} g\left( \tilde{u}_{n + 1}\right)\\
        \implies & u_{n + 1} = u_n + h \left( f(u_{n}, \theta, t_n) - \frac{1}{h} \mathcal{J}_{u}^T \left(\mathcal{J}_{u} \mathcal{J}_{u}^T\right)^{-1} g\left( \tilde{u}_{n + 1}\right) \right)
    \end{aligned}
\end{equation}
Approximating $\frac{1}{h}$ with a fixed term $\gamma$, we recover the Stabilized Neural ODE~\citep{white2023stabilized} formulation.
\begin{equation}
    \frac{\mathrm{d}u}{\mathrm{d}t} = f(u_{n}, \theta, t_n) - \underbrace{\gamma \mathcal{J}_{u}^T \left(\mathcal{J}_{u} \mathcal{J}_{u}^T\right)^{-1} g\left( \tilde{u}_{n + 1}\right)}_{\text{stabilization matrix } F(u)}
\end{equation}
This derivation shows that continuous-relaxation/stabilization strategies can be understood as approximations of our projection-based formulation. In the next section, we empirically evaluate how this formulation translates to improved long-horizon prediction and constraint satisfaction across a range of dynamical systems.

\begin{figure}[t]
    \centering
    \includegraphics[width=\linewidth]{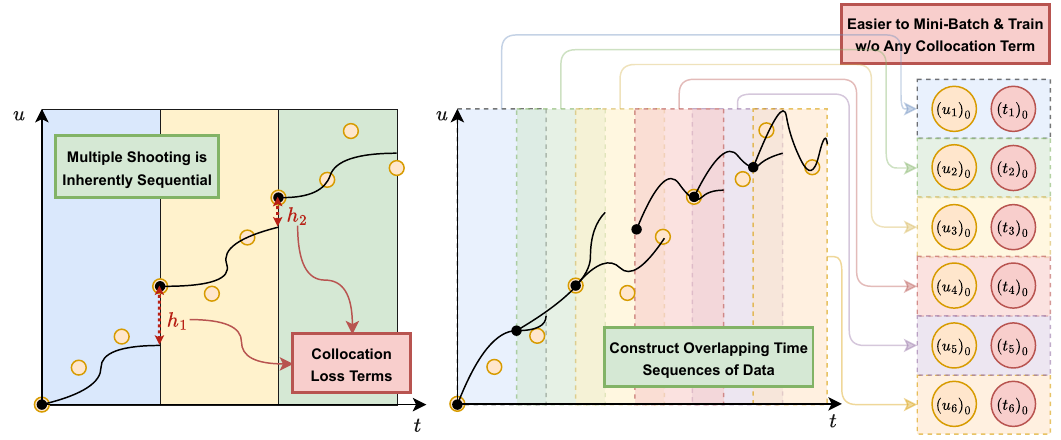}
    \caption{\textbf{Training Pipeline for Neural ODEs:} We train the model using overlapping time sequences, enabling efficient mini-batching without requiring collocation losses. Each batch serves as a short rollout, while continuity across sequences (from overlapping time sequence) allows the model to learn long-horizon dynamics. In contrast, multiple shooting methods optimize trajectory segments sequentially, requiring explicit continuity constraints. Our training setup enables easy batching and avoids the complexity of enforcing inter-segment consistency during optimization.}\label{fig:training-overview}
\end{figure}

\input{dynamics}

\section{Experiments}\label{sec:experiments}

\paragraph{Problem Setup}

We evaluate our models on a suite of constrained dynamical systems [See \Cref{tab:dynamical_systems}]. For each system, we generate training data by simulating trajectories from a fixed set of initial conditions sampled from a known distribution. The model is trained on short time horizons specified per system in \Cref{tab:dynamical_systems}. After training, we evaluate model performance on a separate set of initial conditions drawn from the same distribution. Crucially, these evaluations are conducted over significantly longer time horizons than seen during training, testing the model's ability to extrapolate and remain stable over time.

The benchmark systems~\citep{white2023stabilized,kasim2022constants} include Lotka-Volterra dynamics, mass-spring systems, the two-body problem, nonlinear 2D spring dynamics, a planar robot arm, and rigid body dynamics. Each system exhibits one or more algebraic invariants or constraints that must be preserved. These systems serve to test both the predictive accuracy and constraint-satisfaction ability of each method.

\paragraph{Computational Setup}

All experiments were performed in Julia 1.11.5~\citep{bezanson2012julia}. We use \texttt{Lux.jl} framework~\citep{pal2023efficient, pal2023lux} for the neural network models, with automatic differentiation handled by \texttt{Tracker.jl}~\citep{innes2018fashionable}. \texttt{OrdinaryDiffEq.jl} solver suite~\citep{rackauckas2017differentialequations} is used for forward simulation, and constraint projections were computed using \texttt{NonlinearSolve.jl}~\citep{pal2024nonlinearsolvejl}. All experiments are run on a \texttt{96 Core AMD Ryzen Threadripper PRO 7995WX} with $32$ threads.

\begin{figure}[t]
    \centering
    \includegraphics[width=\linewidth]{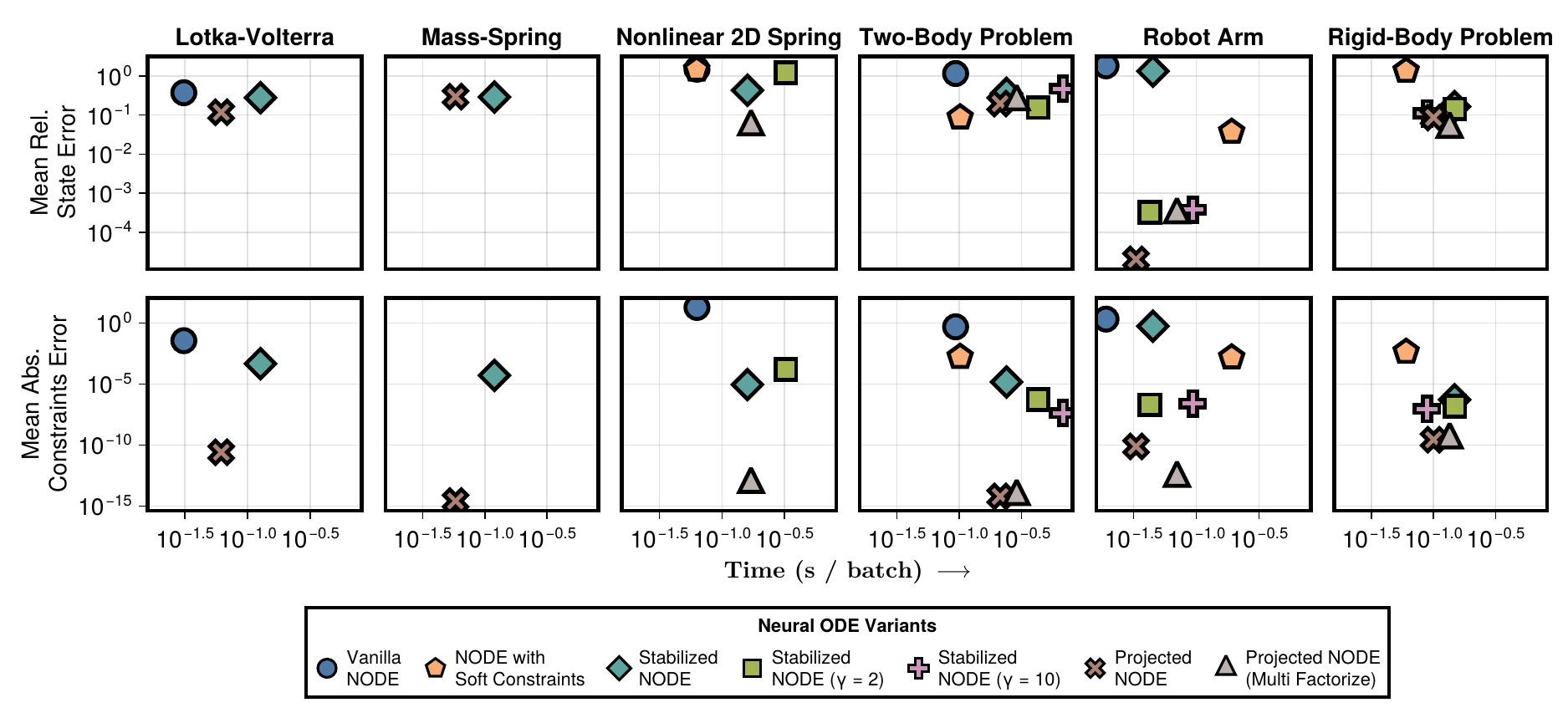}
    \caption{\textbf{Accuracy \& Constraint Satisfaction vs Inference Time}: Each point represents a method's performance on a specific system, showing relative state error (top), constraint violation (bottom), and inference time (x-axis). PNODE variants consistently achieve lower runtime compared to other methods for a given level of mean error.}\label{fig:time_vs_error}
\end{figure}

\begin{table}[t]
    \centering
    \begin{adjustbox}{width=0.95\textwidth}
        \begin{tabular}{Sl Sc Sc Sl Sc Sc Sc}
            \toprule
            \multicolumn{1}{c}{\makecell{\textbf{Dynamical}\\\textbf{System}}} &
            \makecell{\textbf{Train.}\\\textbf{Time}\\\textbf{End}} &
            \makecell{\textbf{Inf.}\\\textbf{Time}\\\textbf{End}} &
            \multicolumn{1}{c}{\makecell{\textbf{Model}}} &
            \multicolumn{1}{c}{\makecell{\textbf{Mean Rel.}\\\textbf{State Error}}} &
            \multicolumn{1}{c}{\makecell{\textbf{Mean Squared}\\\textbf{Constraints Error}}} &
            \multicolumn{1}{c}{\makecell{\textbf{Inf. Time}\\\textbf{(s / batch)}}} \\
            \midrule
            \multirow{3}{*}{Lotka-Volterra} & \multirow{3}{*}{$7.0$} & \multirow{3}{*}{$1000.0$} & NODE & $3.6882 \times 10^{-1} \pm 0.2109$ & $3.6341 \times 10^{-2} \pm 0.0212$ & $0.0310$ \\
             & & & SNODE & $2.7809 \times 10^{-1} \pm 0.3388$ & $4.7076 \times 10^{-4} \pm 0.0005$ & $0.1265$ \\
             & & & \ul{PNODE} & \ul{$1.1782 \times 10^{-1} \pm 0.1436$} & \ul{$2.5989 \times 10^{-11} \pm 0.0000$} & $0.0615$ \\
            \phantom{empty-row} & & & & & & \\

            \multirow{4}{*}{Mass-Spring} & \multirow{4}{*}{$10.0$} & \multirow{4}{*}{$1000.0$} & NODE & \xmark & \xmark & \xmark \\
             & & & NODE + SC & \xmark & \xmark & \xmark \\
             & & & SNODE & \ul{$2.6766 \times 10^{-1} \pm 0.2912$} & $5.1266 \times 10^{-5} \pm 0.0001$ & $0.1192$ \\
             & & & \ul{PNODE} & $2.9497 \times 10^{-1} \pm 0.3330$ & \ul{$2.6347 \times 10^{-15} \pm 0.0000$} & $0.0581$ \\
            \phantom{empty-row} & & & & & & \\

            \multirow{5}{*}{\shortstack[l]{Nonlinear 2D\\Spring}} & \multirow{5}{*}{$10.0$} & \multirow{5}{*}{$1000.0$} & NODE & $1.1861 \times 10^{0} \pm 0.9904$ & $1.9173 \times 10^{1} \pm 347.6511$ & $0.0631$ \\
             & & & NODE + SC & \xmark & \xmark & \xmark \\
             & & & SNODE & $4.3192 \times 10^{-1} \pm 0.5712$ & $9.1205 \times 10^{-6} \pm 0.0000$ & $0.1598$ \\
             & & & SNODE $(\gamma = 2.0)$ & $1.2120 \times 10^{0} \pm 1.0399$ & $1.5555 \times 10^{-4} \pm 0.0007$ & $0.3235$ \\
             & & & \ul{PNODE (MF)} & \ul{$5.0906 \times 10^{-2} \pm 0.0579$} & \ul{$6.4614 \times 10^{-14} \pm 0.0000$} & $0.1705$ \\
            \phantom{empty-row} & & & & & & \\

            \multirow{7}{*}{\shortstack[l]{Two-Body\\Problem}} & \multirow{7}{*}{$6.3832$} & \multirow{7}{*}{$1000.0$} & NODE & $3.5428 \times 10^{0} \pm 2.5194$ & $4.9757 \times 10^{-1} \pm 0.2922$ & $0.0936$ \\
             & & & NODE + SC & $8.7444 \times 10^{-2} \pm 0.1210$ & $1.6729 \times 10^{-3} \pm 0.0004$ & $0.1017$ \\
             & & & SNODE & $3.6016 \times 10^{-1} \pm 0.6181$ & $1.4770 \times 10^{-5} \pm 0.0001$ & $0.2399$ \\
             & & & SNODE $(\gamma = 2.0)$ & $1.5651 \times 10^{-1} \pm 0.4268$ & $5.3088 \times 10^{-7} \pm 0.0000$ & $0.4303$ \\
             & & & SNODE $(\gamma = 10.0)$ & $4.7413 \times 10^{-1} \pm 1.0926$ & $4.3117 \times 10^{-8} \pm 0.0000$  & $0.6830$ \\
             & & & \ul{PNODE} & \ul{$1.9658 \times 10^{-1} \pm 0.3918$} & \ul{$6.2416 \times 10^{-15} \pm 0.0000$} & $0.2133$ \\
             & & & \ul{PNODE (MF)} & $2.2290 \times 10^{-1} \pm 0.4395$ & $6.5457 \times 10^{-15} \pm 0.0000$ & $0.2895$ \\
            \phantom{empty-row} & & & & & & \\

            \multirow{7}{*}{Robot Arm} & \multirow{7}{*}{$5.0$} & \multirow{7}{*}{$250.0$} & NODE & $5.5547 \times 10^{-1} \pm 0.8487$ & $2.1255 \times 10^{0} \pm 6.6763$ & $0.0190$ \\
             & & & NODE + SC & $3.7123 \times 10^{-2} \pm 0.0364$ & $1.5161 \times 10^{-3} \pm 0.0050$ & $0.1916$ \\
             & & & SNODE & $4.5472 \times 10^{-1} \pm 0.7614$ & $5.6041 \times 10^{-1} \pm 1.7733$ & $0.0451$ \\
             & & & SNODE $(\gamma = 2.0)$ & $3.2634 \times 10^{-4} \pm 0.0002$ & $2.0277 \times 10^{-7} \pm 0.0000$ & $0.0427$ \\
             & & & SNODE $(\gamma = 10.0)$ & $3.7928 \times 10^{-4} \pm 0.0004$ & $2.5641 \times 10^{-7} \pm 0.0000$ & $0.0939$ \\
             & & & \ul{PNODE} & \ul{$2.0245 \times 10^{-5} \pm 0.0000$} & $7.7782 \times 10^{-11} \pm 0.0000$ & $0.0330$ \\
             & & & \ul{PNODE (MF)} & $2.8631 \times 10^{-4} \pm 0.0004$ & \ul{$2.0726 \times 10^{-13} \pm 0.0000$} & $0.0701$ \\
            \phantom{empty-row} & & & & & & \\

            \multirow{7}{*}{\shortstack[l]{Rigid-Body\\Problem}} & \multirow{7}{*}{$25.0$} & \multirow{7}{*}{$1000.0$} & NODE & \xmark & \xmark & \xmark \\
             & & & NODE + SC & $1.3492 \times 10^{0} \pm 0.5124$ & $4.3704 \times 10^{-3} \pm 0.0412$ & $0.0603$ \\
             & & & SNODE & $1.6470 \times 10^{-1} \pm 0.2702$ & $5.2363 \times 10^{-7} \pm 0.0000$ & $0.1478$ \\
             & & & SNODE $(\gamma = 2.0)$ & $1.4436 \times 10^{-1} \pm 0.2545$ & $1.5529 \times 10^{-7} \pm 0.0000$ & $0.1482$ \\
             & & & SNODE $(\gamma = 10.0)$ & $1.1091 \times 10^{-1} \pm 0.2027$ & $9.2545 \times 10^{-8} \pm 0.0000$ & $0.0886$ \\
             & & & \ul{PNODE} & $8.6116 \times 10^{-2} \pm 0.1567$ & \ul{$2.7418 \times 10^{-10} \pm 0.0000$} & $0.1001$ \\
             & & & \ul{PNODE (MF)} & \ul{$4.4253 \times 10^{-2} \pm 0.0810$} & $2.9068 \times 10^{-10} \pm 0.0000$ & $0.1347$  \\
            \bottomrule
        \end{tabular}
    \end{adjustbox}
    \par\medskip
    \caption{\textbf{Quantitative performance across benchmark systems.} We report the mean relative state error, mean absolute constraint violation, and inference time (seconds per batch) for each model across all six dynamical systems. PNODEs consistently achieve the lowest constraint violations, often by several orders of magnitude, while maintaining competitive or superior state prediction accuracy. The robust multi-factorization variant (PNODE (MF)) further improves constraint satisfaction in some settings. \xmark{} indicates that the model produced divergent or unstable trajectories during long-horizon evaluation.}\label{tab:results_and_problem_settings}
    \vspace{-2em}
\end{table}

\paragraph{Training Setup}

ODE solver gradients are computed using discrete adjoints via \texttt{SciMLSensitivity.jl}~\citep{ma2021comparison, sapienza2024differentiable}, while the projection step is differentiated using the implicit function theorem~\citep{blondel2022efficient, johnson2012notes}. During training, we construct overlapping time windows from continuous trajectories and treat each window as an independent training sample (see \Cref{fig:training-overview}). This enables efficient batching and avoids the global collocation losses typically required in multiple shooting approaches. We first pretrain on short trajectory segments using the Adam optimizer~\citep{kingma2014adam}, and then fine-tune on full-length trajectories using L-BFGS~\citep{liu1989limited, mogensen2018optim} for improved convergence. All code and experiment scripts are publicly available at \url{https://anonymous.4open.science/r/manifoldneuralodes-neurips/}.

\paragraph{Baseline Models}  

We compare our method against the following baselines:
\begin{enumerate}
    \item \textbf{Neural ODE}: a standard unconstrained formulation.
    \item \textbf{Neural ODE with Soft Constraints}: constraint penalties are added directly to the loss function.
    \item \textbf{Stabilized Neural ODE}~\citep{white2023stabilized}: introduces a stabilization term to keep trajectories near the constraint manifold.
\end{enumerate}

Based on the findings of \citet{gruver2022deconstructing}, we exclude HNNs, as their performance gains are largely attributed to second-order inductive biases rather than structure preservation. For second-order systems such as the mass-spring, nonlinear 2D spring, and two-body problems, we instead use a second-order Neural ODE formulation, which captures these benefits more directly and consistently improves predictive accuracy.

\begin{figure}[t]
    \centering
    \includegraphics[width=\linewidth]{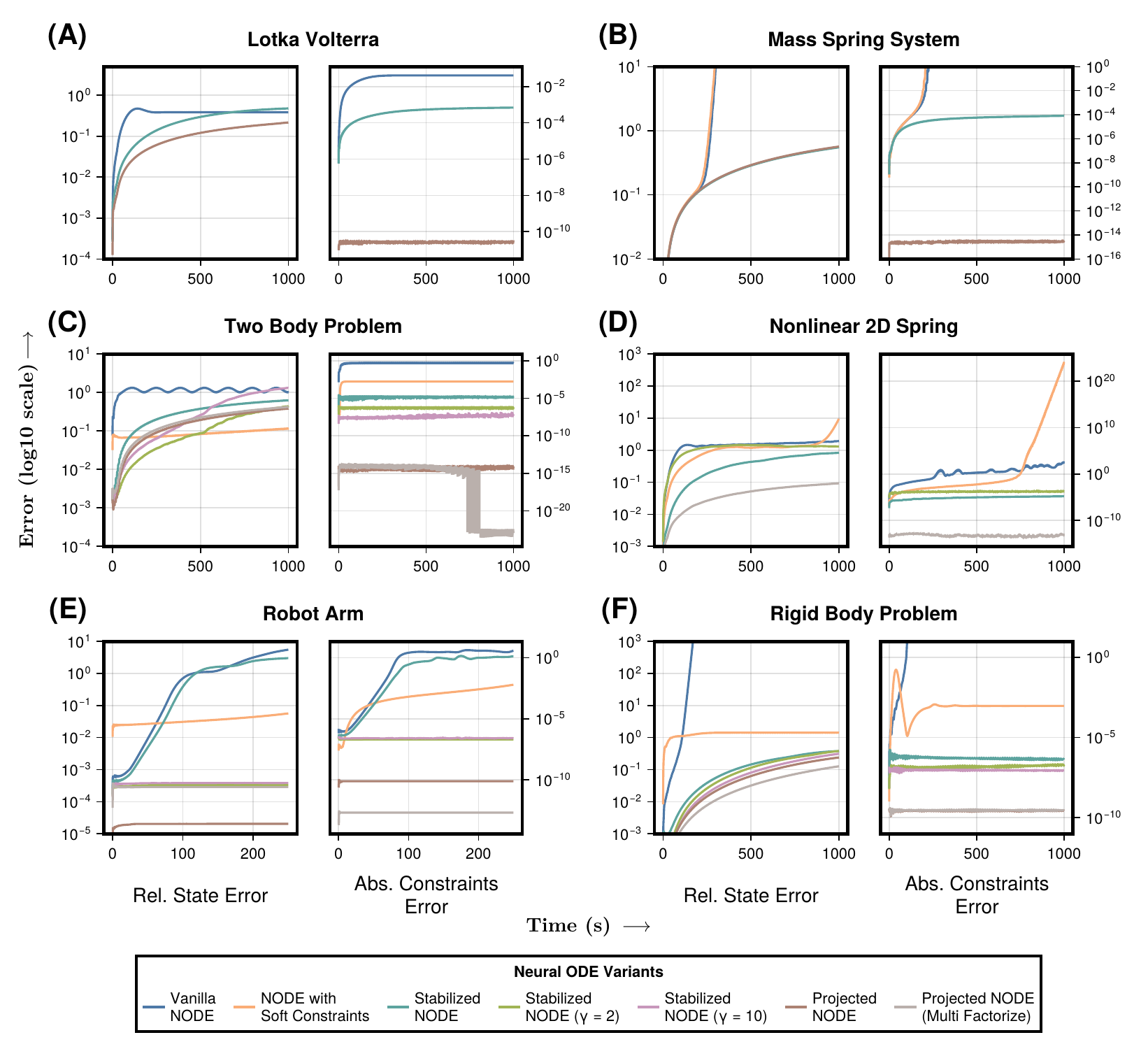}
    \caption{\textbf{Projected Neural ODE has the lowest constraint violations on six benchmark systems.} We report the relative state error (left axis) and absolute constraint violation (right axis) over long integration horizons for each method: Vanilla NODE, NODE with Soft Constraints, Stabilized NODE (with $\gamma$ = $0.5$, $2.0$, $10.0$), and Projected NODEs (with and w/o multiple Jacobian factorizations). Each subplot corresponds to a different dynamical system -- Lotka-Volterra, mass-spring, two-body, nonlinear spring, robot arm, and rigid body. PNODEs consistently maintain several magnitudes lower constraint violations while achieving competitive or superior prediction accuracy.}\label{fig:relative_error_plots}
\end{figure}

\paragraph{Results Summary}

Across all evaluated systems, PNODE variants demonstrate strong performance in both predictive accuracy and constraint satisfaction [See \Cref{tab:results_and_problem_settings}]. In particular, the robust multi-factorization PNODE consistently yields near-zero constraint violation while maintaining state prediction error on par with, or better than, the baselines [See \Cref{fig:relative_error_plots}]. This highlights the effectiveness of explicit manifold projection in preserving the physical structure of constrained systems. Refer to \Cref{fig:motivation-figure}, \Cref{fig:tbp_validation_trajectories}, \Cref{fig:ns2d_validation_trajectories}, \Cref{fig:ra_validation_trajectories}, \& \Cref{fig:rbp_validation_trajectories} for qualitative plots of the model predictions.

The stabilized NODE baselines perform reasonably well on short to medium horizons, especially with tuned values of $\gamma$. However, their performance degrades on longer trajectories, where constraint drift becomes more pronounced. While larger values of $\gamma$ tend to produce better constraint satisfaction, the dynamics become stiff which warrant the use of slower implicit ODE solver. Moreover, the stabilization comes at a significant computational cost, as reflected in longer inference times. In contrast, PNODEs offer a better accuracy–efficiency tradeoff [See \Cref{fig:time_vs_error}], especially when using the single-factorization variant. Soft-constraint approaches improve constraint satisfaction over vanilla NODEs but remain unstable in certain systems (e.g., Mass-Spring, Rigid Body), where even small violations accumulate rapidly. In some cases, such as the Nonlinear 2D Spring, they lead to divergent trajectories.

Notably, the inference efficiency of PNODEs is a key strength. For a fixed level of prediction error, PNODEs are often faster than stabilized NODEs, and even comparable to vanilla NODEs [See \Cref{fig:time_vs_error}]. This makes PNODEs suitable not just for accuracy-critical applications, but also for deployment in real-time or resource-constrained settings.

\section{Conclusion \& Discussion}\label{sec:conclusion}

We introduced Projected Neural ODEs (PNODEs), a principled framework for learning continuous-time dynamics under algebraic constraints. By projecting each ODE step onto the constraint manifold, PNODEs enforce physical invariants explicitly during training and inference. We derived two variants -- one robust and iterative, the other fast and approximate -- both grounded in numerical methods for semi-explicit DAEs. Our framework also unifies and generalizes several recent stabilization approaches.

Empirical results across a diverse suite of dynamical systems show that PNODEs achieve significantly better long-horizon accuracy and constraint satisfaction than existing baselines. The multi-factorization variant is especially robust across all problems, while the single-factorization variant offers competitive speed. In contrast, soft-penalty methods and stabilization approaches suffer from either constraint drift, inference overhead, or sensitivity to hyperparameters. Our results demonstrate that manifold projection is a simple yet powerful design principle for learning constrained dynamics.

\paragraph{Limitations}

Our study focuses exclusively on semi-explicit DAEs. While many real-world systems can be modeled in this form, not all DAEs admit a semi-explicit structure. Extending PNODEs to handle more general implicit DAE formulations remains an important direction for future work. Additionally, our experiments assume a fixed, non-parameterized constraint manifold. Although PNODEs are not inherently limited to fixed manifolds, further empirical validation is needed to assess performance on systems with parameterized or time-varying constraint manifolds.

\section*{Acknowledgments}

This material is based upon work supported by the U.S. National Science Foundation under award Nos CNS-2346520, PHY-2028125, RISE-2425761, DMS-2325184, OAC-2103804, and OSI-2029670, by the Defense Advanced Research Projects Agency (DARPA) under Agreement No. HR00112490488, by the Department of Energy, National Nuclear Security Administration under Award Number DE-NA0003965 and by the United States Air Force Research Laboratory under Cooperative Agreement Number FA8750-19-2-1000. Neither the United States Government nor any agency thereof, nor any of their employees, makes any warranty, express or implied, or assumes any legal liability or responsibility for the accuracy, completeness, or usefulness of any information, apparatus, product, or process disclosed, or represents that its use would not infringe privately owned rights. Reference herein to any specific commercial product, process, or service by trade name, trademark, manufacturer, or otherwise does not necessarily constitute or imply its endorsement, recommendation, or favoring by the United States Government or any agency thereof. The views and opinions of authors expressed herein do not necessarily state or reflect those of the United States Government or any agency thereof." The views and conclusions contained in this document are those of the authors and should not be interpreted as representing the official policies, either expressed or implied, of the United States Air Force or the U.S. Government.

We thank Frank Sch\"afer for early discussions on the feasibility of projection-based neural ODEs, Vaibhav Dixit for insights on training strategies, and Khushi Doshi for helpful feedback on figure design and overall presentation of the paper.

\bibliography{ref}
\bibliographystyle{icml2024}

\newpage
\appendix

\listoftodos

\input{appendix}

\end{document}

%% file: dynamics.tex
\begin{table}[t]
    \centering
    \begin{adjustbox}{width=0.95\textwidth}
        \begin{tabular}{Sl Sc Sc}
            \toprule
            \textbf{Dynamical System} & \textbf{ODE System} & \textbf{Algebraic Invariant(s)} \\
            \midrule
            \textit{Lotka-Volterra} &
            \makecell[l]{$
                \begin{aligned}
                    \frac{\mathrm{d}x}{\mathrm{d}t} &= \alpha x - \beta x y \\
                    \frac{\mathrm{d}y}{\mathrm{d}t} &= -\gamma y + \delta x y
                \end{aligned}
            $} &
            \makecell[l]{$
                V(x, y) = \delta x - \gamma \ln{x} + \beta y - \alpha \ln{y}
            $} \\
            \phantom{empty-row} & & \\
            \textit{Mass Spring} &
            \makecell[l]{$
                \begin{aligned}
                    \frac{\mathrm{d}x}{\mathrm{d}t} = v \qquad
                    \frac{\mathrm{d}v}{\mathrm{d}t} = -x
                \end{aligned}
            $} &
            \makecell[l]{$
                E(v, x) = \frac{1}{2} \left(x^2 + v^2\right)
            $} \\
            \phantom{empty-row} & & \\
            \textit{Two-Body Problem} &
            \makecell[l]{$
                \begin{aligned}
                    \frac{\mathrm{d}q_1}{\mathrm{d}t} &= p_1 \qquad \frac{\mathrm{d}p_1}{\mathrm{d}t} = \frac{-q_1}{\left(q_1^2 + q_2^2\right)^{\frac{3}{2}}} \\
                    \frac{\mathrm{d}q_2}{\mathrm{d}t} &= p_2 \qquad \frac{\mathrm{d}p_2}{\mathrm{d}t} = \frac{-q_2}{\left(q_1^2 + q_2^2\right)^{\frac{3}{2}}}
                \end{aligned}
            $} &
            \makecell[l]{$
                E(q_1, q_2, p_1, p_2) = q_1 p_2 - q_2 p_1
            $} \\
            \phantom{empty-row} & & \\
            \textit{Nonlinear 2D Spring} &
            \makecell[l]{$
                \begin{aligned}
                    \frac{\mathrm{d}x}{\mathrm{d}t} &= u \qquad \frac{\mathrm{d}u}{\mathrm{d}t} = -x \left(x^2 + y^2\right) \\
                    \frac{\mathrm{d}y}{\mathrm{d}t} &= v \qquad \frac{\mathrm{d}v}{\mathrm{d}t} = -y \left(x^2 + y^2\right)
                \end{aligned}
            $} & 
            \makecell[l]{$
                \begin{aligned}
                    E(x, y, u, v) =& \frac{1}{2} \left(u^2 + v^2\right) + \\
                                   & \frac{1}{4} \left(x^2 + y^2\right)^2 \\
                    L(x, y, u, v) =& x v - y u
                \end{aligned}
            $} \\
            \phantom{empty-row} & & \\
            \textit{Robot Arm} &
            \makecell[l]{$
                \begin{aligned}
                    e(\theta) &= \begin{bmatrix}
                        \cos(\theta_1) + \cos(\theta_2) + \cos(\theta_3)\\
                        \sin(\theta_1) + \sin(\theta_2) + \sin(\theta_3)
                    \end{bmatrix}\\
                    p(t) &= e_0 - \begin{bmatrix}
                        \frac{\sin\left(2\pi t\right)}{2\pi}\\
                        0
                    \end{bmatrix} \\
                    \dot{\theta} &= e^\prime(\theta)^T \left(e^\prime(\theta) e^\prime(\theta)^T\right)^{-1} \dot{p}(t)
                \end{aligned}
            $} & 
            \makecell[l]{$
                \begin{aligned}
                    C(\theta, t) = e(\theta) - p(t)
                \end{aligned}
            $} \\
            \phantom{empty-row} & & \\
            \textit{Rigid Body Problem} &
            \makecell[l]{$
                \begin{aligned}
                    \frac{\mathrm{d}y_1}{\mathrm{d}t} &= \left(\frac{1}{I_3} - \frac{1}{I_2}\right) y_2 y_3 \\
                    \frac{\mathrm{d}y_2}{\mathrm{d}t} &= \left(\frac{1}{I_1} - \frac{1}{I_3}\right) y_1 y_3 \\
                    \frac{\mathrm{d}y_3}{\mathrm{d}t} &= \left(\frac{1}{I_2} - \frac{1}{I_1}\right) y_1 y_2
                \end{aligned}
            $} & 
            \makecell[l]{$
                \begin{aligned}
                    C(y_1, y_2, y_3) &= \frac{1}{2} \left(y_1^2 + y_2^2 + y_3^2\right)
                \end{aligned}
            $} \\
            \bottomrule
        \end{tabular}
    \end{adjustbox}
    \par\medskip
    \caption{\textbf{Summary of benchmark dynamical systems and their algebraic invariants.} We evaluate PNODEs on a diverse set of dynamical systems that exhibit known algebraic constraints or conserved quantities. Each system is defined by its governing ODEs and associated invariants, including energy, angular momentum, or kinematic constraints. These benchmarks test the model's ability to preserve system-specific structure over long horizons.}\label{tab:dynamical_systems}
\end{table}

%% file: appendix.tex
\newpage
\section{Qualitative Visualization of Learned Dynamical Systems}

\begin{figure}[H]
    \centering
    \includegraphics[width=\linewidth]{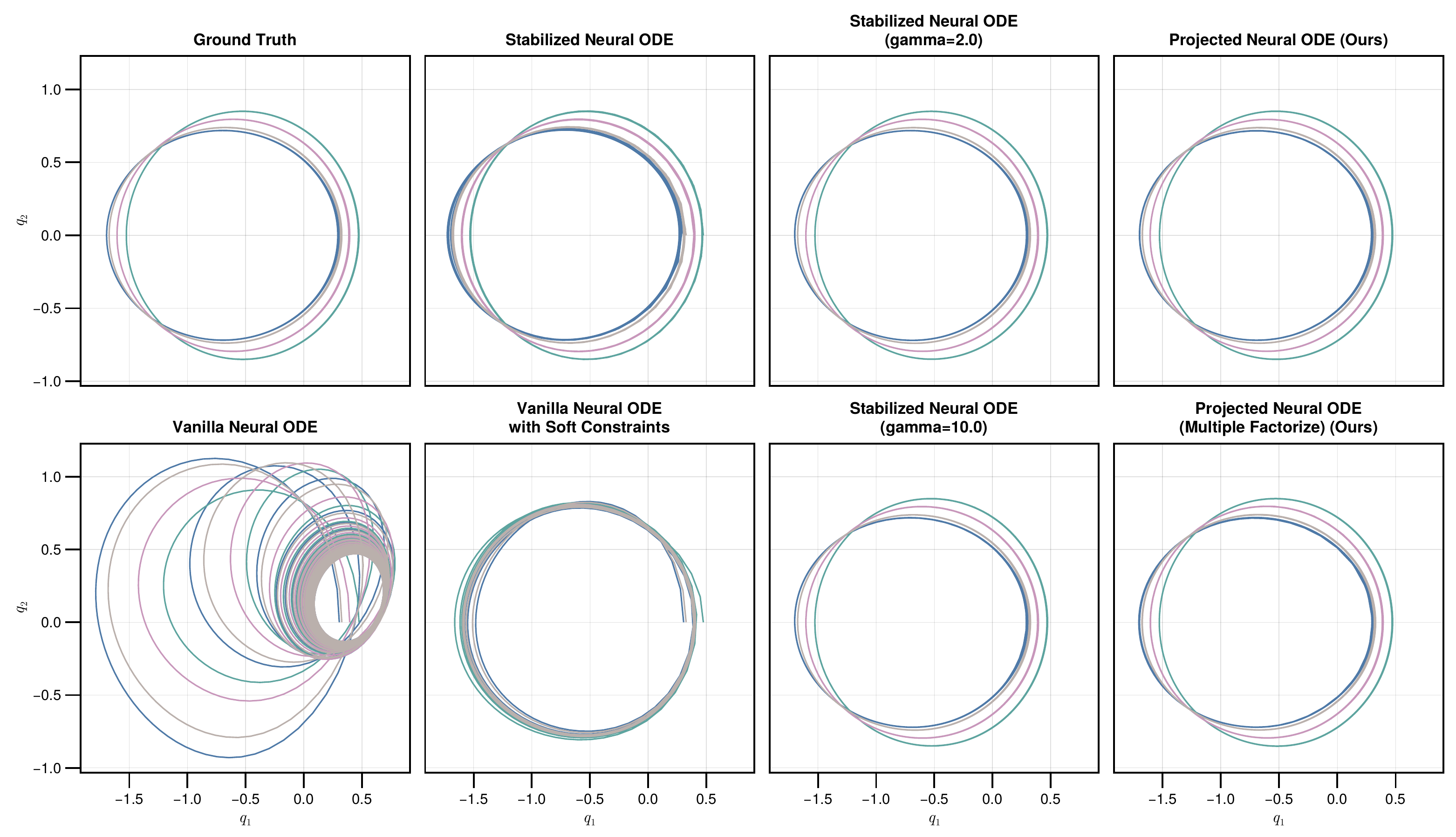}
    \caption{\textbf{Qualitative Plots of Predicted Trajectories for the Two-Body problem}}\label{fig:tbp_validation_trajectories}
\end{figure}

\begin{figure}[H]
    \centering
    \includegraphics[width=\linewidth]{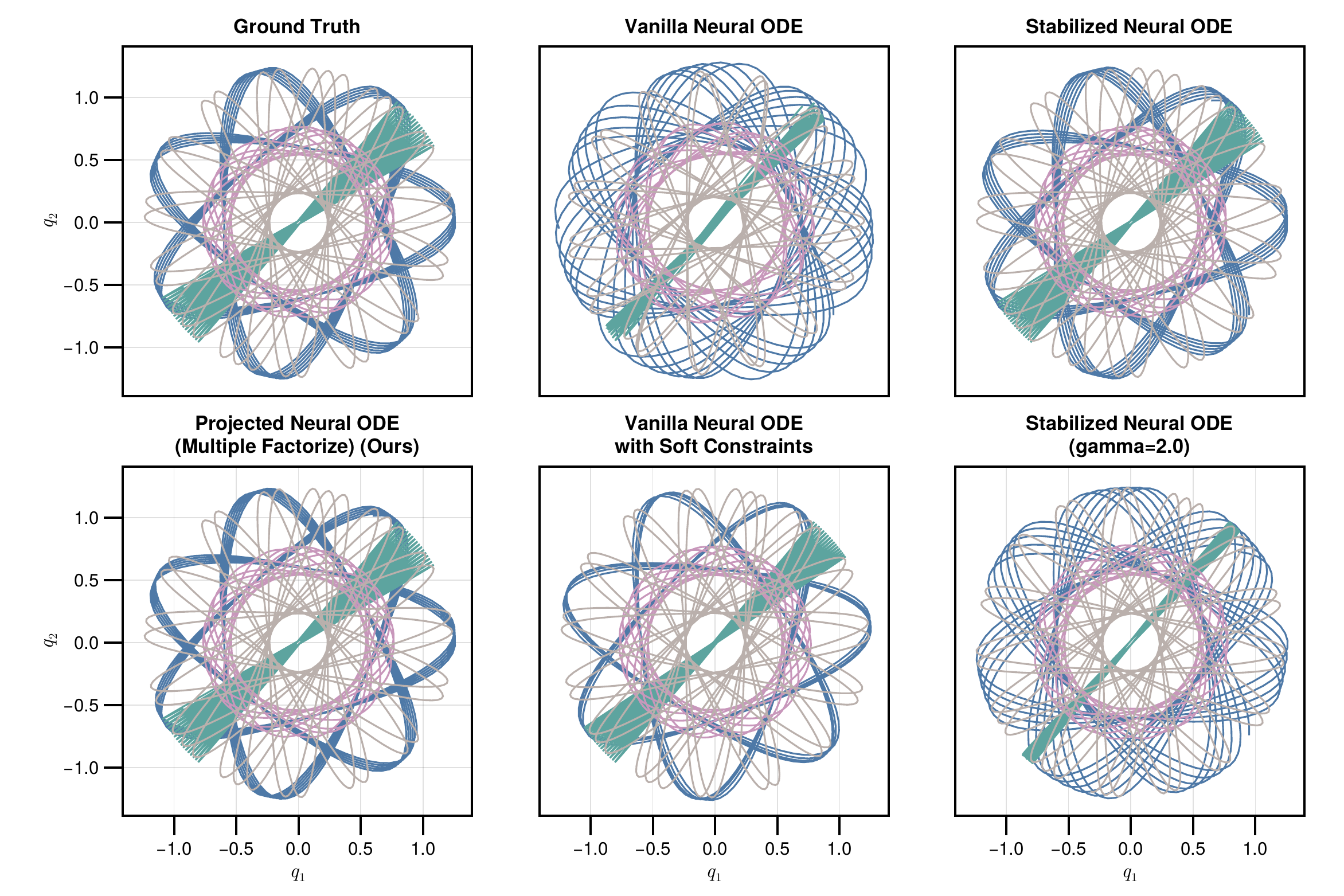}
    \caption{\textbf{Qualitative Plots of Predicted Trajectories for the Nonlinear Spring 2D problem.}}\label{fig:ns2d_validation_trajectories}
\end{figure}

\begin{figure}[H]
    \centering
    \includegraphics[width=\linewidth]{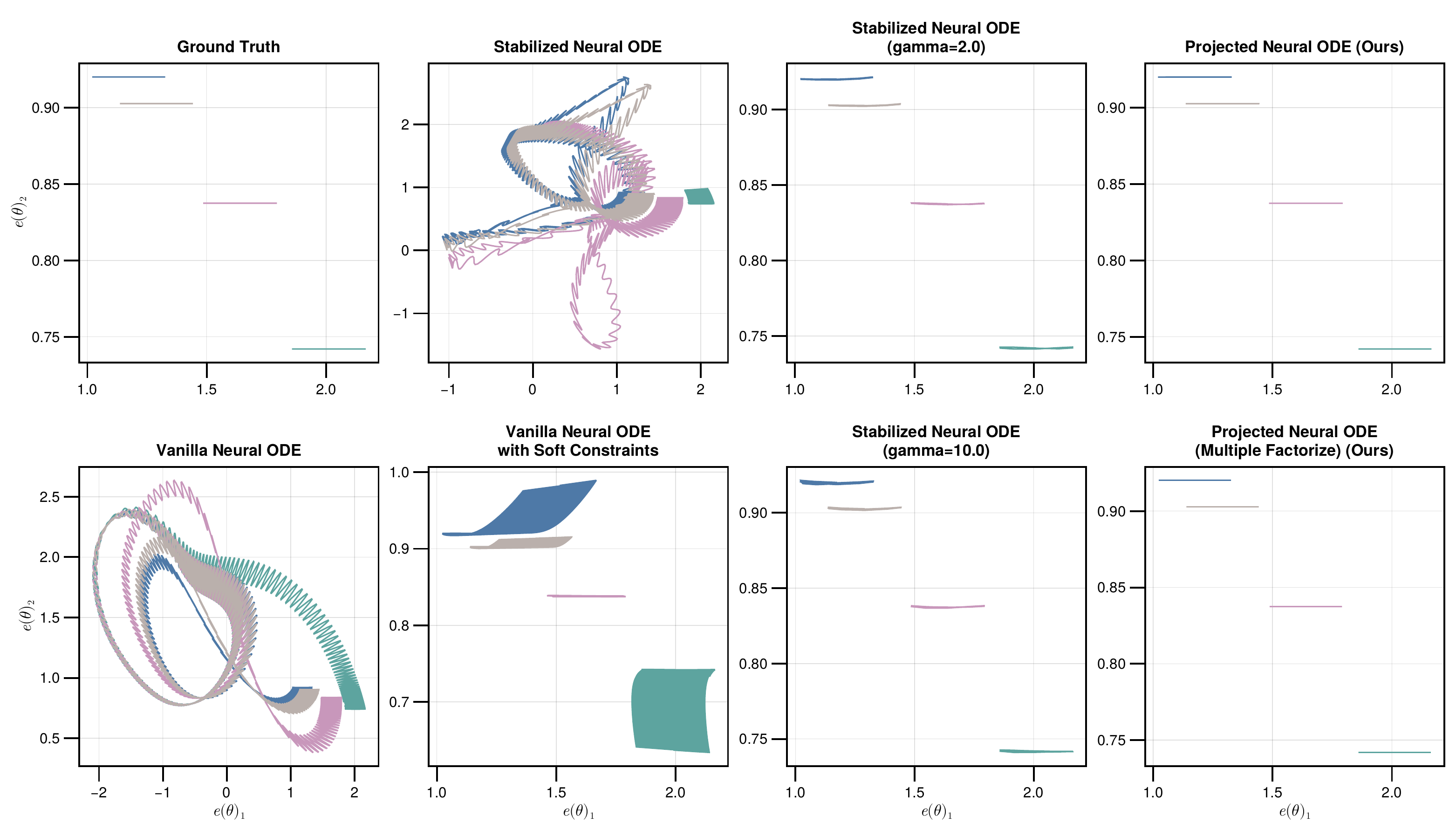}
    \caption{\textbf{Qualitative Plots of Predicted Trajectories for the Robot Arm problem.}}\label{fig:ra_validation_trajectories}
\end{figure}

\begin{figure}[H]
    \centering
    \includegraphics[width=\linewidth]{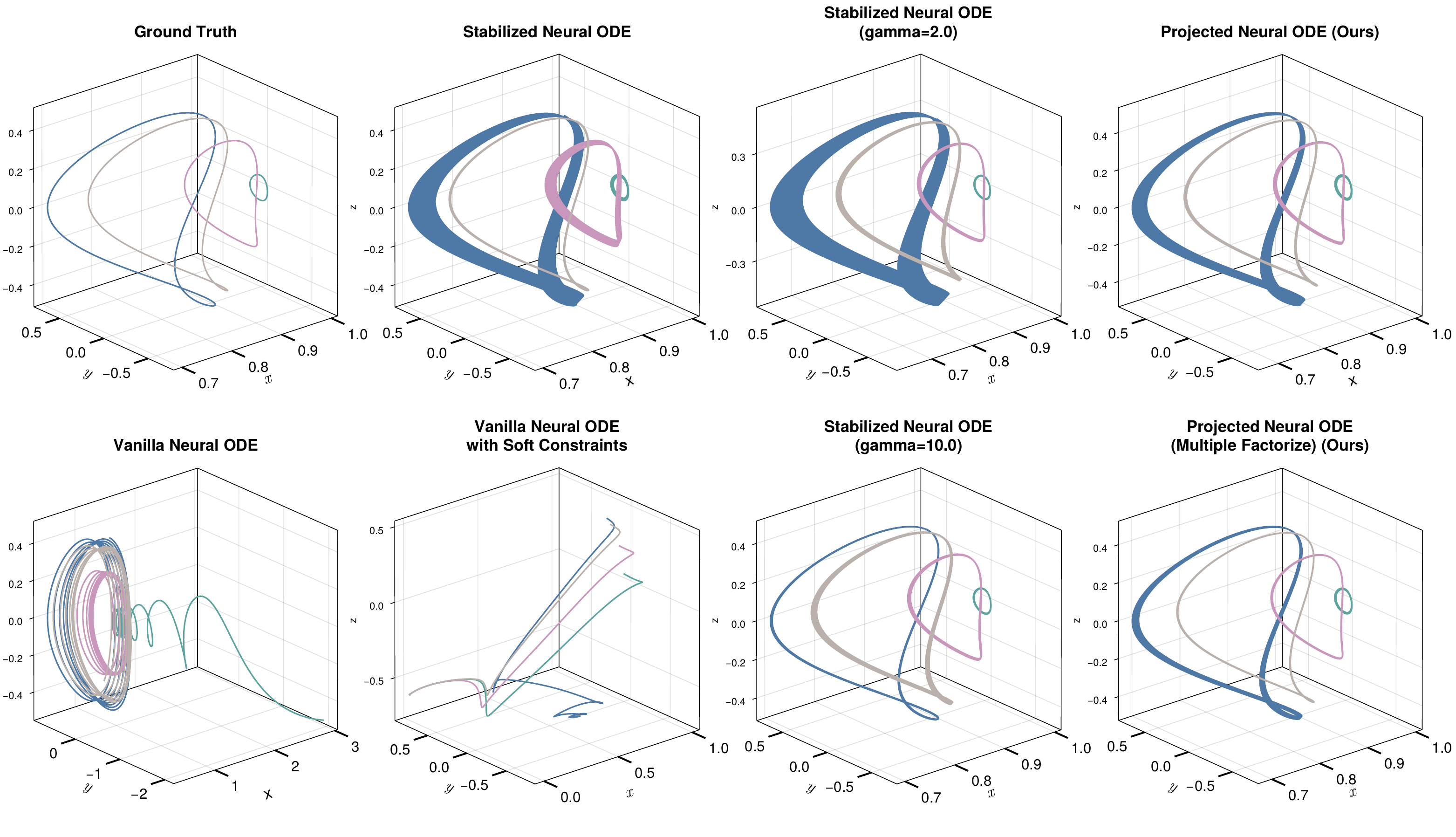}
    \caption{\textbf{Qualitative Plots of Predicted Trajectories for the Rigid-Body problem.} The vanilla NODE case produces divergent trajectory and is plotted for a shorter time-horizon compared to the other cases.}\label{fig:rbp_validation_trajectories}
\end{figure}

%% file: main.bbl
\begin{thebibliography}{35}
\providecommand{\natexlab}[1]{#1}
\providecommand{\url}[1]{\texttt{#1}}
\expandafter\ifx\csname urlstyle\endcsname\relax
  \providecommand{\doi}[1]{doi: #1}\else
  \providecommand{\doi}{doi: \begingroup \urlstyle{rm}\Url}\fi

\bibitem[Bezanson et~al.(2012)Bezanson, Karpinski, Shah, and
  Edelman]{bezanson2012julia}
Bezanson, J., Karpinski, S., Shah, V.~B., and Edelman, A.
\newblock Julia: A fast dynamic language for technical computing.
\newblock \emph{arXiv preprint arXiv:1209.5145}, 2012.

\bibitem[Blondel et~al.(2022)Blondel, Berthet, Cuturi, Frostig, Hoyer,
  Llinares-L{\'o}pez, Pedregosa, and Vert]{blondel2022efficient}
Blondel, M., Berthet, Q., Cuturi, M., Frostig, R., Hoyer, S.,
  Llinares-L{\'o}pez, F., Pedregosa, F., and Vert, J.-P.
\newblock Efficient and modular implicit differentiation.
\newblock \emph{Advances in neural information processing systems},
  35:\penalty0 5230--5242, 2022.

\bibitem[Chen et~al.(2018)Chen, Rubanova, Bettencourt, and
  Duvenaud]{chen2018neural}
Chen, R.~T., Rubanova, Y., Bettencourt, J., and Duvenaud, D.~K.
\newblock Neural ordinary differential equations.
\newblock \emph{Advances in neural information processing systems}, 31, 2018.

\bibitem[Cohen et~al.(1996)Cohen, Hindmarsh, Dubois, et~al.]{cohen1996cvode}
Cohen, S.~D., Hindmarsh, A.~C., Dubois, P.~F., et~al.
\newblock Cvode, a stiff/nonstiff ode solver in c.
\newblock \emph{Computers in physics}, 10\penalty0 (2):\penalty0 138--143,
  1996.

\bibitem[Cranmer et~al.(2020)Cranmer, Greydanus, Hoyer, Battaglia, Spergel, and
  Ho]{cranmer2020lagrangian}
Cranmer, M., Greydanus, S., Hoyer, S., Battaglia, P., Spergel, D., and Ho, S.
\newblock Lagrangian neural networks.
\newblock \emph{arXiv preprint arXiv:2003.04630}, 2020.

\bibitem[David \& M{\'e}hats(2023)David and M{\'e}hats]{david2023symplectic}
David, M. and M{\'e}hats, F.
\newblock Symplectic learning for hamiltonian neural networks.
\newblock \emph{Journal of Computational Physics}, 494:\penalty0 112495, 2023.

\bibitem[Greydanus et~al.(2019)Greydanus, Dzamba, and
  Yosinski]{greydanus2019hamiltonian}
Greydanus, S., Dzamba, M., and Yosinski, J.
\newblock Hamiltonian neural networks.
\newblock \emph{Advances in neural information processing systems}, 32, 2019.

\bibitem[Griepentrog \& M{\"a}rz(1986)Griepentrog and
  M{\"a}rz]{griepentrog1986differential}
Griepentrog, E. and M{\"a}rz, R.
\newblock Differential-algebraic equations and their numerical treatment.
\newblock \emph{(No Title)}, 1986.

\bibitem[Gruver et~al.(2022)Gruver, Finzi, Stanton, and
  Wilson]{gruver2022deconstructing}
Gruver, N., Finzi, M., Stanton, S., and Wilson, A.~G.
\newblock Deconstructing the inductive biases of hamiltonian neural networks.
\newblock \emph{arXiv preprint arXiv:2202.04836}, 2022.

\bibitem[Hairer(2011)]{hairer2011solving}
Hairer, E.
\newblock Solving differential equations on manifolds.
\newblock \emph{Lecture notes}, 2011.

\bibitem[Hairer et~al.(2006{\natexlab{a}})Hairer, Hochbruck, Iserles, and
  Lubich]{hairer2006geometric}
Hairer, E., Hochbruck, M., Iserles, A., and Lubich, C.
\newblock Geometric numerical integration.
\newblock \emph{Oberwolfach Reports}, 3\penalty0 (1):\penalty0 805--882,
  2006{\natexlab{a}}.

\bibitem[Hairer et~al.(2006{\natexlab{b}})Hairer, Lubich, and
  Roche]{hairer2006numerical}
Hairer, E., Lubich, C., and Roche, M.
\newblock \emph{The numerical solution of differential-algebraic systems by
  Runge-Kutta methods}, volume 1409.
\newblock Springer, 2006{\natexlab{b}}.

\bibitem[Innes et~al.(2018)Innes, Saba, Fischer, Gandhi, Rudilosso, Joy,
  Karmali, Pal, and Shah]{innes2018fashionable}
Innes, M., Saba, E., Fischer, K., Gandhi, D., Rudilosso, M.~C., Joy, N.~M.,
  Karmali, T., Pal, A., and Shah, V.
\newblock Fashionable modelling with flux, 2018.

\bibitem[Johnson(2012)]{johnson2012notes}
Johnson, S.~G.
\newblock Notes on adjoint methods for 18.335.
\newblock \emph{Introduction to Numerical Methods}, 2012.

\bibitem[Karniadakis et~al.(2021)Karniadakis, Kevrekidis, Lu, Perdikaris, Wang,
  and Yang]{karniadakis2021physics}
Karniadakis, G.~E., Kevrekidis, I.~G., Lu, L., Perdikaris, P., Wang, S., and
  Yang, L.
\newblock Physics-informed machine learning.
\newblock \emph{Nature Reviews Physics}, 3\penalty0 (6):\penalty0 422--440,
  2021.

\bibitem[Kasim \& Lim(2022)Kasim and Lim]{kasim2022constants}
Kasim, M.~F. and Lim, Y.~H.
\newblock Constants of motion network.
\newblock \emph{Advances in Neural Information Processing Systems},
  35:\penalty0 25295--25305, 2022.

\bibitem[Kidger(2022)]{kidger2022neural}
Kidger, P.
\newblock On neural differential equations.
\newblock \emph{arXiv preprint arXiv:2202.02435}, 2022.

\bibitem[Kingma(2014)]{kingma2014adam}
Kingma, D.~P.
\newblock Adam: A method for stochastic optimization.
\newblock \emph{arXiv preprint arXiv:1412.6980}, 2014.

\bibitem[Koch et~al.(2024)Koch, Shapiro, Sharma, Vrabie, and
  Drgona]{koch2024neural}
Koch, J., Shapiro, M., Sharma, H., Vrabie, D., and Drgona, J.
\newblock Neural differential algebraic equations.
\newblock \emph{arXiv preprint arXiv:2403.12938}, 2024.

\bibitem[Lim \& Kasim(2022)Lim and Kasim]{lim2022unifying}
Lim, Y.~H. and Kasim, M.~F.
\newblock Unifying physical systems' inductive biases in neural ode using
  dynamics constraints.
\newblock \emph{arXiv preprint arXiv:2208.02632}, 2022.

\bibitem[Liu \& Nocedal(1989)Liu and Nocedal]{liu1989limited}
Liu, D.~C. and Nocedal, J.
\newblock On the limited memory bfgs method for large scale optimization.
\newblock \emph{Mathematical programming}, 45\penalty0 (1):\penalty0 503--528,
  1989.

\bibitem[Ma et~al.(2021)Ma, Dixit, Innes, Guo, and
  Rackauckas]{ma2021comparison}
Ma, Y., Dixit, V., Innes, M.~J., Guo, X., and Rackauckas, C.
\newblock A comparison of automatic differentiation and continuous sensitivity
  analysis for derivatives of differential equation solutions.
\newblock In \emph{2021 IEEE High Performance Extreme Computing Conference
  (HPEC)}, pp.\  1--9. IEEE, 2021.

\bibitem[Mattheakis et~al.(2022)Mattheakis, Sondak, Dogra, and
  Protopapas]{mattheakis2022hamiltonian}
Mattheakis, M., Sondak, D., Dogra, A.~S., and Protopapas, P.
\newblock Hamiltonian neural networks for solving equations of motion.
\newblock \emph{Physical Review E}, 105\penalty0 (6):\penalty0 065305, 2022.

\bibitem[Mogensen \& Riseth(2018)Mogensen and Riseth]{mogensen2018optim}
Mogensen, P. and Riseth, A.
\newblock Optim: A mathematical optimization package for julia.
\newblock \emph{Journal of Open Source Software}, 3\penalty0 (24), 2018.

\bibitem[Pal(2023{\natexlab{a}})]{pal2023efficient}
Pal, A.
\newblock {On Efficient Training \& Inference of Neural Differential
  Equations}, 2023{\natexlab{a}}.

\bibitem[Pal(2023{\natexlab{b}})]{pal2023lux}
Pal, A.
\newblock {Lux: Explicit Parameterization of Deep Neural Networks in Julia},
  April 2023{\natexlab{b}}.
\newblock URL \url{https://doi.org/10.5281/zenodo.7808903}.
\newblock If you use this software, please cite it as below.

\bibitem[Pal et~al.(2024)Pal, Holtorf, Larsson, Loman, Utkarsh, Schaefer, Qu,
  Edelman, and Rackauckas]{pal2024nonlinearsolvejl}
Pal, A., Holtorf, F., Larsson, A., Loman, T., Utkarsh, Schaefer, F., Qu, Q.,
  Edelman, A., and Rackauckas, C.
\newblock {NonlinearSolve.jl: High-Performance and Robust Solvers for Systems
  of Nonlinear Equations in Julia}, 2024.

\bibitem[Rackauckas \& Nie(2017)Rackauckas and
  Nie]{rackauckas2017differentialequations}
Rackauckas, C. and Nie, Q.
\newblock Differentialequations. jl--a performant and feature-rich ecosystem
  for solving differential equations in julia.
\newblock \emph{Journal of open research software}, 5\penalty0 (1):\penalty0
  15--15, 2017.

\bibitem[Rackauckas et~al.(2020)Rackauckas, Ma, Martensen, Warner, Zubov,
  Supekar, Skinner, Ramadhan, and Edelman]{rackauckas2020universal}
Rackauckas, C., Ma, Y., Martensen, J., Warner, C., Zubov, K., Supekar, R.,
  Skinner, D., Ramadhan, A., and Edelman, A.
\newblock Universal differential equations for scientific machine learning.
\newblock \emph{arXiv preprint arXiv:2001.04385}, 2020.

\bibitem[Raissi et~al.(2019)Raissi, Perdikaris, and
  Karniadakis]{raissi2019physics}
Raissi, M., Perdikaris, P., and Karniadakis, G.~E.
\newblock Physics-informed neural networks: A deep learning framework for
  solving forward and inverse problems involving nonlinear partial differential
  equations.
\newblock \emph{Journal of Computational physics}, 378:\penalty0 686--707,
  2019.

\bibitem[Rathore et~al.(2024)Rathore, Lei, Frangella, Lu, and
  Udell]{rathore2024challenges}
Rathore, P., Lei, W., Frangella, Z., Lu, L., and Udell, M.
\newblock Challenges in training pinns: A loss landscape perspective.
\newblock \emph{arXiv preprint arXiv:2402.01868}, 2024.

\bibitem[Sapienza et~al.(2024)Sapienza, Bolibar, Sch{\"a}fer, Groenke, Pal,
  Boussange, Heimbach, Hooker, P{\'e}rez, Persson,
  et~al.]{sapienza2024differentiable}
Sapienza, F., Bolibar, J., Sch{\"a}fer, F., Groenke, B., Pal, A., Boussange,
  V., Heimbach, P., Hooker, G., P{\'e}rez, F., Persson, P.-O., et~al.
\newblock Differentiable programming for differential equations: A review.
\newblock \emph{arXiv preprint arXiv:2406.09699}, 2024.

\bibitem[Von~Schwerin(1999)]{von1999multibody}
Von~Schwerin, R.
\newblock \emph{Multibody system simulation: numerical methods, algorithms, and
  software}, volume~7.
\newblock Springer Science \& Business Media, 1999.

\bibitem[White et~al.(2023)White, Kilbertus, Gelbrecht, and
  Boers]{white2023stabilized}
White, A., Kilbertus, N., Gelbrecht, M., and Boers, N.
\newblock Stabilized neural differential equations for learning constrained
  dynamics.
\newblock \emph{arXiv preprint arXiv:2306.09739}, 2023.

\bibitem[White et~al.(2024)White, Büttner, Gelbrecht, Duruisseaux, Kilbertus,
  Hellmann, and Boers]{white2024projectedneuraldifferentialequations}
White, A., Büttner, A., Gelbrecht, M., Duruisseaux, V., Kilbertus, N.,
  Hellmann, F., and Boers, N.
\newblock Projected neural differential equations for learning constrained
  dynamics, 2024.
\newblock URL \url{https://arxiv.org/abs/2410.23667}.

\end{thebibliography}
